%% file: manuscript.tex
\documentclass[a4paper,fleqn]{cas-dc}

\usepackage[numbers]{natbib}

\def\tsc#1{\csdef{#1}{\textsc{\lowercase{#1}}\xspace}}
\tsc{WGM}
\tsc{QE}
\tsc{EP}
\tsc{PMS}
\tsc{BEC}
\tsc{DE}

\input{setting}

\begin{document}
\let\WriteBookmarks\relax
\def\floatpagepagefraction{0.7}
\def\textpagefraction{.001}
\shorttitle{VoxelDiffusionCut}
\shortauthors{T. Hachimine et~al.}

\title [mode = title]{
VoxelDiffusionCut: Non-destructive Internal-part Extraction via Iterative Cutting and Structure Estimation
}

\author[1]{Takumi Hachimine\orcidlink{0009-0006-0180-4168}}[orcid=0009-0006-0180-4168]

\author[1]{Yuhwan Kwon\orcidlink{0000-0001-9058-1379}}[orcid=0000-0001-9058-1379]

\author[1]{Cheng-Yu Kuo\orcidlink{0000-0002-3085-0343}}[orcid=0000-0002-3085-0343]

\author[1]{Tomoya Yamanokuchi\orcidlink{0000-0003-2387-2197}}[orcid=0000-0003-2387-2197]
\cormark[1]
\ead{yamanokuchi.tomoya@naist.ac.jp}

\author[1]{Takamitsu Matsubara\orcidlink{0000-0003-3545-4814}}[orcid=0000-0003-3545-4814]

\affiliation[1]{organization={Division of Information Science, Graduate School of Science and Technology, Nara Institute of Science and Technology},
    addressline={8916-5 Takayama}, 
    city={Ikoma}, 
    state={Nara}, 
    postcode={630-0192}, 
    country={Japan}
}

\cortext[cor1]{Corresponding author}

\begin{abstract}
    Non-destructive extraction of the target internal part, such as batteries and motors, by cutting surrounding structures is crucial at recycling and disposal sites.
    However, the diversity of products and the lack of information on disassembly procedures make it challenging to decide where to cut.
    This study explores a method for non-destructive extraction of a target internal part that iteratively estimates the internal structure from observed cutting surfaces and formulates cutting plans based on the estimation results.
    A key requirement is to estimate the probability of the target part's presence from partial observations.
    However, learning conditional generative models for this task is challenging: The high dimensionality of 3D shape representations makes learning difficult, and conventional models (e.g., conditional variational autoencoders) often fail to capture multi-modal predictive uncertainty due to mode collapse, resulting in overconfident predictions.
    To address these issues, we propose VoxelDiffusionCut, which iteratively estimates the internal structure represented as voxels using a diffusion model and plans cuts for non-destructive extraction of the target internal part based on the estimation results.
    Voxel representation allows the model to predict only attributes at fixed grid positions, i.e., types of constituent parts, making learning more tractable.
    The diffusion model completes the voxel representation conditioned on observed cutting surfaces, capturing uncertainty in unobserved regions to avoid erroneous cuts.
    Experimental results in simulation suggest that the proposed method can estimate internal structures from observed cutting surfaces and enable non-destructive extraction of the target internal part by leveraging the estimated uncertainty.
\end{abstract}


\maketitle

\section{Introduction}
    Extracting internal parts within products is an essential task at recycling and disposal sites \cite{KANTADAS2024221,UEDA2024139928}.
    These parts, such as batteries, motors, and electronic components, must be extracted without damage to ensure safety and their functionality.
    If product information is available, internal parts can be extracted through disassembly \cite{Kiyokawa2025dis,9419740,9492060,8653968}.
    However, such information is rarely disclosed due to concerns about intellectual property, and disassembly is often infeasible due to adhesives or degraded joints.
    Even when products share the same external shape, their internal structures may vary due to minor modifications introduced in different production years or specifications, making part extraction difficult.
    
    \begin{figure}
        \centering
        \vspace{3truemm}
        \includegraphics[clip, width=0.8\linewidth]{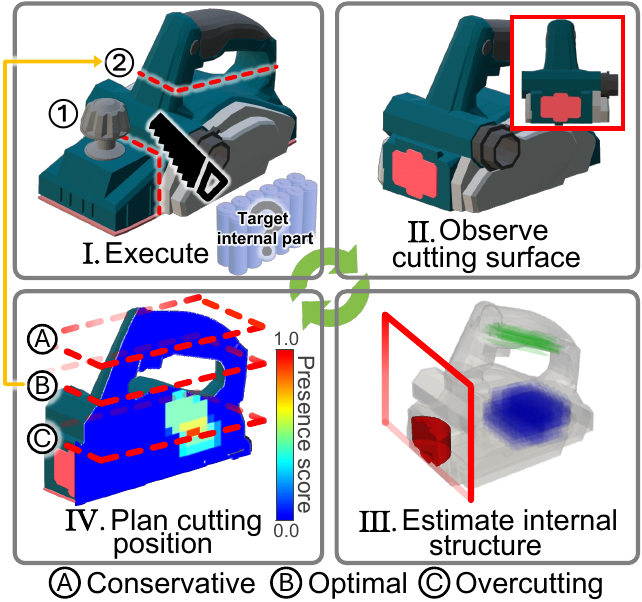}
        \caption{
        Overview of the process for non-destructive extraction of the target internal part.
        The process iteratively estimates the internal structure from the observed cutting surfaces and plans cutting positions to extract the target part without damage.
        }
        \vspace{-3.5truemm}
        \label{fig:eye_catch}
    \end{figure}

    Mechanical separation, such as \textit{cutting}, is a practical method applicable even for products with adhesives or degradation \cite{9966530}; however, deciding where to cut remains challenging due to the unknown internal structure.
    Therefore, we explore a method for non-destructive extraction of the target internal part from a product with an unknown internal structure, based on cutting surfaces observed during the cutting process (\figurename\ref{fig:eye_catch}).
    As a first step toward such automation, this study focuses on the extraction of a target internal part from products with identical external shapes but different internal structures. To focus on the fundamental challenges inherent in the problem, we adopt an idealized setting without modeling observation noise or cutting errors.\footnote{Practical extensions are discussed in \S\ref{sec:discussion}.}

    A key difficulty in non-destructive extraction of the target internal part lies in estimating the entire internal structure from partial observations on cutting surfaces.
    Therefore, we consider an iterative approach:
    (a) estimating the internal structure based on observed cutting surfaces, and
    (b) planning cutting positions that maximize removable volume while avoiding damage to the target internal part.
    To achieve non-destructive extraction, it is necessary to estimate the probability of the target part's presence from partial observations, which entails estimating the uncertainty of the internal structure in the unobserved regions.
    This estimation can be viewed as a conditional generative modeling problem, where internal structures are generated conditioned on partial observations.
    However, learning such models poses two challenges: (1) Directly learning internal structures as 3D shapes (e.g., point clouds or meshes) is difficult due to the high dimensionality of these representations and their unordered data structures \cite{qi2017pointnet,wang2019dynamic,hanocka2019meshcnn}, which makes it hard to reproduce internal structures sufficiently and can lead to erroneous cuts.
    (2) Conventional generative models, such as Conditional Variational Autoencoders (CVAEs) \cite{NIPS2015_8d55a249} and Conditional Generative Adversarial Networks (CGANs) \cite{mirza2014conditional} often suffer from failure to capture multi-modal predictive uncertainty due to mode collapse, resulting in overconfident predictions.

    To address these issues, we view internal structure estimation as 3D shape completion in \textit{voxel space}, where the observed cutting surfaces provide partial observations of the voxelized shape.
    By representing 3D shapes as voxels and encoding constituent parts as voxel attributes, the model only needs to predict these attributes at fixed grid positions. This regular grid structure can make learning more tractable compared to directly handling irregular 3D geometry.
    We formulate this voxel completion problem using recent generative models, such as diffusion models \cite{sohl2015deep,ho2020denoising,dhariwal2021diffusion,lugmayr2022repaint}, conditioned on the observed cutting surfaces.
    Diffusion models flexibly capture multi-modal predictive uncertainty in unobserved regions through iterative denoising-based generation, thereby mitigating mode collapse and alleviating overconfident predictions, which helps reduce the risk of incorrect cutting plans.

    With the above in mind, we propose VoxelDiffusionCut, a framework for non-destructive extraction of a target internal part from a product with an unknown internal structure, using observed cutting surfaces.
    VoxelDiffusionCut consists of iterative internal structure estimation in a voxel representation using a diffusion model and cutting planning.
    The estimation results are incorporated in the cutting plan to maximize removable volume while avoiding damage to the target internal part.

    We evaluated the proposed method in a simulator using simple-shaped models and complex-shaped models that emulate real products.
    The experimental results suggest that our method can estimate internal structures from observed cutting surfaces while capturing uncertainty in unobserved regions.
    Moreover, cutting plans based on these estimations enable non-destructive extraction of the target internal part.

    The key contributions of this paper are as follows:
    \begin{itemize}
      \item We formulate a new problem setting for product dismantling, in which a target internal part is extracted by cutting a product with an unknown internal structure.
      \item We propose VoxelDiffusionCut, a novel framework for non-destructive extraction of a target internal part from a product, based on iterative voxel-based internal structure estimation using a diffusion model and cutting planning.
      \item We validate the proposed VoxelDiffusionCut through simulation experiments using simple- and complex-shaped models, where the complex-shaped models emulate real products.
    \end{itemize}

\section{Related Work}
    \subsection{Studies on Dismantling Automation for Non-Destructive Internal-part Extraction}
    Many studies have addressed parts extraction at recycling and disposal sites \cite{MISHRA2024100900,9966530,10155458}.
    If product information is available, disassembly can be used to extract internal parts \cite{Kiyokawa2025dis,9419740,9492060,8653968}.
    However, such information is rarely disclosed due to concerns about intellectual property.
    Therefore, product dismantling approaches based on mechanical shredding
    using shredders or rotating chains, combined with part sorting, have
    been investigated \cite{LU20221,9551431}.
    Tsunazawa et al. showed that optimizing chain rotation speed and crushing time enables efficient recovery of waste printed circuit boards (PCBs) \cite{TSUNAZAWA2018474}.
    Such approaches are effective for material recovery but unsuitable when valuable parts must not be shredded.

    To extract parts that are difficult to shred, methods using X-ray imaging, including tomographic reconstruction techniques, have been proposed to visualize internal structures \cite{LI201988} or identify parts for separation \cite{STERKENS2021105246}.
    Ueda et al. proposed extracting internal parts by applying impact at the product joints \cite{UEDA2024139928}. The optimal impact position is selected using a genetic algorithm based on the battery location obtained from X-ray imaging.
    However, such X-ray-based approaches assume that internal structures are clearly captured by X-ray equipment, which restricts their applicability to thin products, such as smartphones. For thick products, such as vacuum cleaners, X-ray imaging often lacks sufficient penetration depth to provide detailed internal information.

    In contrast, mechanical separation by cutting can be applied regardless of product thickness, but the internal structure remains unknown, which makes the decision of where to cut challenging.
    Therefore, we explore a new framework for the extraction of the target internal part that combines internal structure estimation by a diffusion model and iterative cutting planning.
    This approach can be applied even to thick products, since internal structure estimation relies on cutting surfaces obtained through cutting.
    \subsection{Diffusion Models for 3D Shape Generation}
    Diffusion models \cite{sohl2015deep,ho2020denoising,dhariwal2021diffusion} are a class of generative models with high expressiveness and flexibility.
    They have been applied to diverse tasks, such as text-to-image generation \cite{ramesh2022hierarchical,saharia2022photorealistic,zhao2023uni}, video editing \cite{pmlr-v235-cohen24a,Kara_2024_CVPR,zhang2023towards} and robot control \cite{janner2022diffuser,mishra2023generative,Ze2024DP3}.

    As an example of 3D shape generation, Luo et al. proposed a diffusion model for point cloud generation \cite{9578791}.
    Tang et al. proposed a method that generates 3D shapes from depth images using two diffusion models trained in a hierarchical manner \cite{NEURIPS2022_40e56dab}.
    Nichol et al. reported a method for generating diverse 3D shapes conditioned on text \cite{nichol2022point}, which combines a text-to-image model with a 3D shape-generation model.
    In addition, diffusion models have been applied to other 3D shape representations, including explicit representations (voxel grids \cite{10656999} and meshes \cite{liu2023meshdiffusion}) and implicit representations (occupancy fields \cite{10658125} and neural radiance fields \cite{wynn2023diffusionerf}).
    However, these approaches are designed to generate 3D object geometry and do not explicitly represent internal structures composed of multiple parts.
    Moreover, directly learning 3D shapes is challenging due to the high dimensionality of shape representations \cite{qi2017pointnet,wang2019dynamic,hanocka2019meshcnn}.

    In contrast, our method represents 3D shapes as voxels and encodes constituent parts as
    voxel attributes, enabling internal structure representation.
    With this shape representation, the model only needs to predict these attributes on a regular grid structure, which makes the learning more tractable.
    Furthermore, we complete the internal structure using a diffusion model conditioned on the observed cutting surfaces.
    This approach captures multi-modal predictive uncertainty in unobserved regions, thereby mitigating mode collapse and overconfident predictions.
    By leveraging the estimated uncertainty during cutting planning, we can help to achieve non-destructive extraction of the target internal part.

\section{Preliminary}
As preparation for the proposed method, we describe diffusion models (\S\ref{subsection:preliminary_ddpm}) and conditional diffusion models (\S\ref{subsection:preliminary_conditional_ddpm}). Specifically, we introduce Classifier-Free Guidance (CFG) generation \cite{ho2021classifierfree,10203721} as a conditional diffusion model for completing partial observations.
\subsection{Denoising Diffusion Probabilistic Models}\label{subsection:preliminary_ddpm}
Diffusion models \cite{sohl2015deep,ho2020denoising,dhariwal2021diffusion} consist of a forward diffusion process, which gradually transforms the original data into Gaussian noise, and a denoising process, which transforms Gaussian noise into samples from the data distribution.
Let $\mathbf{x}^{0}$ denote the original data and $\mathbf{x}^{n}$ the data after adding noise $n$ times.
The forward diffusion process is defined as
$q(\mathbf{x}^{1:N}|\mathbf{x}^{0})\coloneqq\prod_{n=1}^{N}q(\mathbf{x}^{n}|\mathbf{x}^{n-1})$, $q(\mathbf{x}^{n}|\mathbf{x}^{n-1})\coloneqq\mathcal{N}(\mathbf{x}^{n};\sqrt{{1-\beta_{n}}}\mathbf{x}^{n-1},\beta_{n}\mathbf{I}).$
Here, $\beta_{n}$ is the noise scheduler, which controls the noise scale.

The denoising process is represented as a Gaussian distribution, where the mean is given by a model parameterized by $\theta$, which takes the noise data $\mathbf{x}^n$ and the diffusion step $n$ as inputs:
\begin{align}
p_{\theta}(\mathbf{x}^{0:N})&\coloneqq p(\mathbf{x}^{N}) \prod_{n=1}^{N}p_{\theta}(\mathbf{x}^{n-1}|\mathbf{x}^{n}),\\
p_{\theta}(\mathbf{x}^{n-1}|\mathbf{x}^n)&\coloneqq \mathcal{N}(\mathbf{x}^{n-1};\boldsymbol{\mu}_{\theta}(\mathbf{x}^{n},n),\mathbf{\Sigma}^n), \label{eq:ddpm_sampleing_step}\\
p(\mathbf{x}^{N})&=\mathcal{N}(\mathbf{0},\mathbf{I}).
\end{align}
The covariance matrix $\mathbf{\Sigma}^n$ is given by $\mathbf{\Sigma}^n = \tilde{\beta}_{n}\mathbf{I}$ \cite{ho2020denoising}, where $\tilde{\beta}_{n} = \beta_{n}(1-\bar{\alpha}_{n-1})/(1-\bar{\alpha}_{n})$, $\bar{\alpha}_{n} = \prod_{s=1}^{n}\alpha_{s}$, and $\alpha_{s} = 1 - \beta_{s}$.
In the training process, instead of directly optimizing $\boldsymbol{\mu}_{\theta}$, the model learns to predict the noise $\epsilon$ from noisy data $\mathbf{x}^{n}$
using the simplified loss function:
\begin{equation}
\mathcal{L}_{\theta} = \mathbb{E}_{n,\epsilon,\mathbf{x}^{0}}\left[\|\epsilon - \epsilon_{\theta}(\mathbf{x}^{n},n)\|^2\right],
\end{equation}
since $\boldsymbol{\mu}_{\theta}$ can be expressed in terms of $\epsilon_{\theta}$ as
$\boldsymbol{\mu}_{\theta}(\mathbf{x}^{n},n) = \frac{1}{\sqrt{\alpha_{n}}}(\mathbf{x}^{n} - \frac{1-\alpha_{n}}{\sqrt{1-\alpha_{n}}}\epsilon_{\theta}(\mathbf{x}^{n},n))$.
\subsection{Conditional Diffusion with Classifier-Free Guidance}\label{subsection:preliminary_conditional_ddpm}
Conditional diffusion models \cite{rombach2022high,ho2021classifierfree} generate data $\mathbf{x}^0$ based on conditions $\mathbf{c}$ such as labels, text, or observed information.
They are trained by minimizing the conditional loss:
\begin{equation}
\mathcal{L}_{\theta}^{\text{cond}} = \mathbb{E}_{n, \epsilon, \mathbf{x}^{0}, \mathbf{c}} \left[ \| \epsilon - \epsilon_{\theta}(\mathbf{x}^{n}, n, \mathbf{c}) \|^2 \right].
\end{equation}

To strengthen conditioning during sampling, classifier guidance \cite{dhariwal2021diffusion} has been proposed, which steers the denoising process using gradients from an external classifier.
However, classifier guidance requires training an additional classifier and may become unreliable when the conditioning information is weak or ambiguous.

To mitigate this, Classifier-Free Guidance (CFG) \cite{ho2021classifierfree} has been proposed.
CFG enables a single diffusion model to perform both conditional and unconditional generation by randomly dropping the condition $\mathbf{c}$ during training.
This prevents overfitting to the condition and allows flexible, diverse generation.
During the denoising process, the predicted noise is computed as a linear combination of the conditional $\epsilon_{\theta}(\mathbf{x}^{n}, n, \mathbf{c})$ and unconditional $\epsilon_{\theta}(\mathbf{x}^{n}, n)$ predictions:
\begin{equation}
\epsilon_{\text{guided}} = (1 + w)\epsilon_{\theta}(\mathbf{x}^{n}, n, \mathbf{c}) - w\epsilon_{\theta}(\mathbf{x}^{n}, n). \label{eq:cfg_epsilon_guidance}
\end{equation}
Here, $w$ is the guidance scale that controls the conditioning strength.

\section{Proposed Method}
\begin{figure*}
        \centering
        \includegraphics[clip, width=1.0 \linewidth]{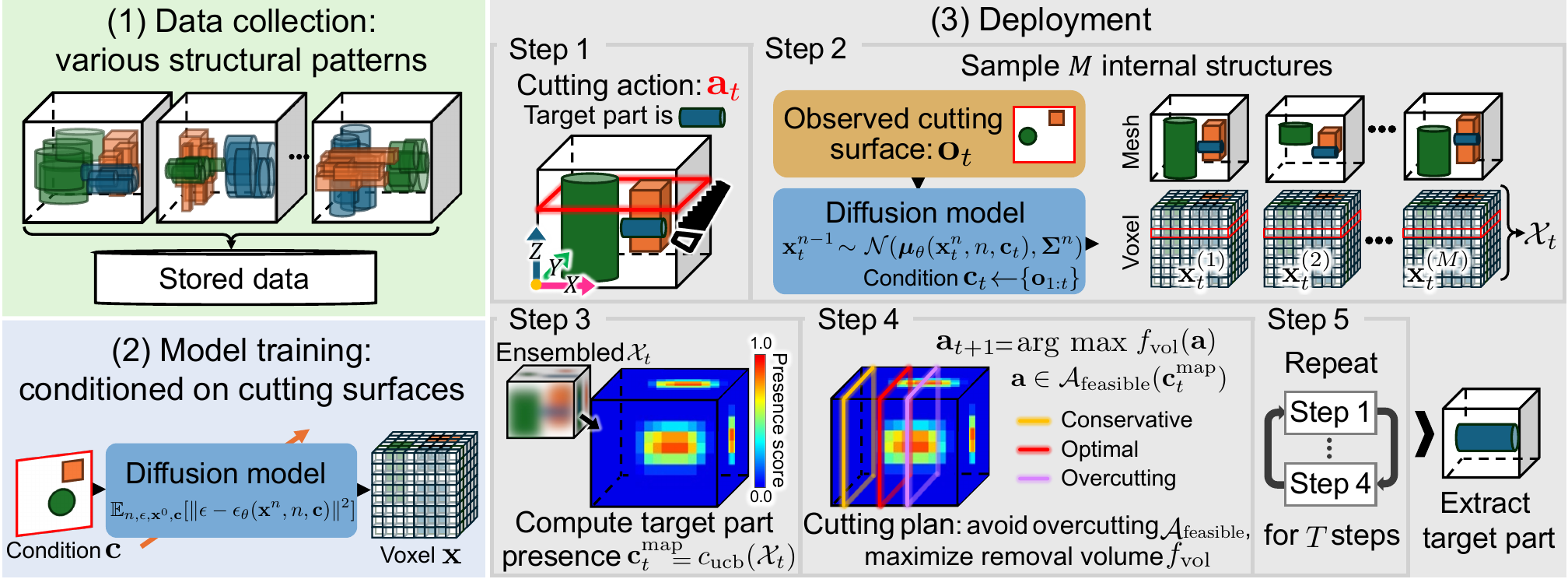}
        \caption{Framework of VoxelDiffusionCut.
         The internal structure prediction model is trained using a diffusion model based on data with various location patterns of internal parts.
         During deployment, the model estimates the internal structure through conditional generation based on the observed cutting surfaces and subsequently plans the cutting positions.}
        \label{fig:proposed_method_overview}
\end{figure*}

In this section, we describe the proposed VoxelDiffusionCut framework
for non-destructive extraction of the target internal part from a product with an unknown internal structure (\S\ref{subsection:proposed_framework}).
\S\ref{subsection:3d_2d_mapping} introduces voxel-based 3D shape representation, \S\ref{subsection:ddpim_internal_structure_prediction} describes conditional diffusion models for internal-structure estimation, and \S\ref{subsection:cutting_planning} presents the cutting action planning method for extracting the target internal part without damage.
Video~1 (00:26--01:37) provides an illustrative explanation of the framework components and the iterative inference--planning loop shown in \figurename\ref{fig:proposed_method_overview}.
\subsection{VoxelDiffusionCut Framework}\label{subsection:proposed_framework}
VoxelDiffusionCut consists of three main stages (\figurename\ref{fig:proposed_method_overview}):

(1) \textbf{Data collection} Collect 3D shapes with multiple internal-structure arrangements and types of parts as training data for an internal-structure prediction model that uses a diffusion model.

(2) \textbf{Model training} Voxelize the collected 3D shapes and encode the constituent parts as voxel attributes.
Then train the diffusion model, conditioned on the cutting surfaces, to estimate the internal structure within the voxel space.

(3) \textbf{Deployment} Extract the target internal part without damaging it by performing the following steps:

\textbf{Step~1}. Execute the cutting action $\mathbf{a}_{t}\in \mathcal{A}$,
where $t \in \{1,\dots,T\}$ denotes the task step.
The initial cutting action is set to avoid damaging the target part, assuming that prior information, such as the product type, is available.
\textbf{Step~2}. Predict the internal structures $\mathbf{x}_{t}$ using the diffusion model conditioned on the observed cutting surface $\mathbf{o}_{t}$.
Diffusion models generate data by sampling from a learned distribution, which leads to diverse outputs. Such sample-based diversity has been used as an indicator of predictive uncertainty in robotic applications~\cite{11127844,11127730,NEURIPS2023_4c5722ba}.
Following this approach, we also leverage this sample-based uncertainty to inform our cutting plan. Therefore, we predict $M$ internal structures $\mathcal{X}_{t} = \{ \mathbf{x}_{t}^{(m)} \}_{m=1}^{M}$.
\textbf{Step~3}. Compute the presence score map $\mathbf{c}_{t}^{\text{map}}$ of the target part using ${c}_{\text{ucb}}(\mathcal{X}_{t})$.
\textbf{Step~4}. Plan the next cutting action $\mathbf{a}_{t+1}$ based on $\mathbf{c}_{t}^{\text{map}}$ to maximize removable volume while avoiding damage to the target part.
\textbf{Step~5}. Repeat Steps~1--4 for $T$ task steps, which enables VoxelDiffusionCut to achieve non-destructive extraction of the target internal part.
\subsection{3D Shape Representation by Voxelization with Part-Attribute Encoding}\label{subsection:3d_2d_mapping}
We represent 3D shapes as voxels and simplify the training of the internal-structure prediction model by learning the voxel attributes that encode the constituent parts.
In the voxel representation, the 3D space is divided along each axis into $K$ uniform cubes, which gives a total of $K^3$ voxels.
Each voxel is assigned attributes (e.g., color, material type) to represent constituent parts of the 3D shape.
We denote $\mathbf{p}_k$ as the 3D coordinates of the $k$-th voxel and $\mathbf{f}_k$ as the feature vector representing its attributes.
A 3D shape is then represented in the voxel space as
$\mathcal{V} = \{\mathbf{v}_{k}\} = \{(\mathbf{p}_{k}, \mathbf{f}_{k})\}.$

\begin{figure}
\begin{minipage}[t]{\columnwidth}
    \centering
        \includegraphics[clip,width=1.0\columnwidth]{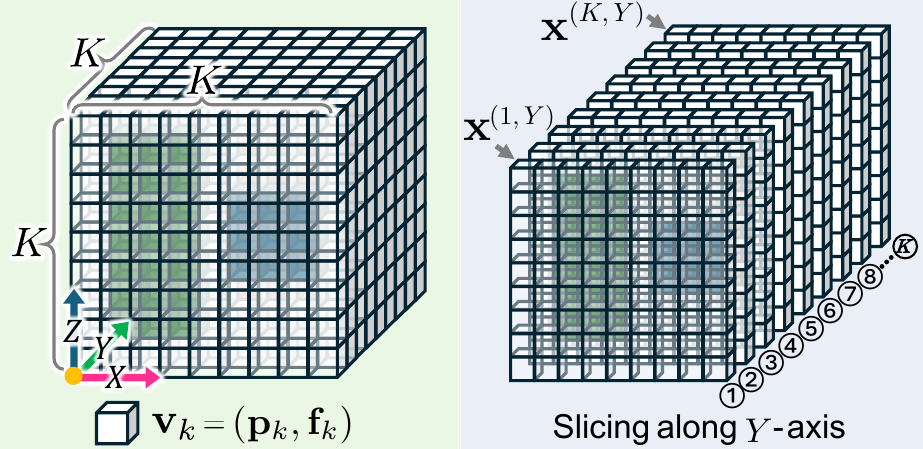}
    \begin{minipage}[b]{0.45\columnwidth}
        \centering
        \vspace{2truemm}
        \subcaption{Voxel representation}
        \label{fig:vox_img_convert:3d_shape}
    \end{minipage}
    \begin{minipage}[b]{0.45\columnwidth}
        \centering
        \vspace{2truemm}
        \subcaption{Cutting surface examples}
        \label{fig:vox_img_convert:3d_voxel}
    \end{minipage}
    \caption{
            3D shape representation by voxelization.
            (a) Voxel representation with a discrete 3D grid \(K\).
            (b) Examples of cutting surfaces (slices) extracted along the \(Y\)-axis (indexed by \(j\)), with each surface denoted as \(\mathbf{x}^{(i,j)}\).
            }
    \label{fig:vox_img_convet}
\end{minipage}
\end{figure}

Therefore, the observation $\mathbf{o}_{t}$ obtained by cutting can be interpreted as the cutting surface of the internal structure $\mathbf{x}_{t}^{(i,j)}$ ($\mathbf{o}_{t} \coloneqq \mathbf{x}_{t}^{(i,j)}$).
Here, $j \in \{X, Y, Z\}$ is the cutting axis and $i \in \{1, \dots, K\}$ specifies the cutting position.
The cutting action can be represented as $\mathbf{a}_{t} \coloneqq [i, j]$.
Figure~\ref{fig:vox_img_convert:3d_shape} shows a 3D voxel representation,
and \figurename\ref{fig:vox_img_convert:3d_voxel} illustrates cutting-surface examples obtained by the cutting action.

\subsection{Internal Structure Estimation Using Conditional Diffusion Models}\label{subsection:ddpim_internal_structure_prediction}
We use a conditional diffusion model with CFG to estimate the internal structure $\mathbf{x}_{t} \in \mathbb{R}^{K \times K \times K \times 3}$ given observations up to task step $t$, denoted as $\mathbf{x}_{t}^{\prime} \coloneqq \{\mathbf{o}_{1:t}\}$.
We define $\mathbf{x}_{t}^{\prime}$ and its mask matrix $\mathbf{m}_{t}\in \mathbb{Z}_{2}^{K\times K \times K}$ as condition $\mathbf{c}_{t}\coloneqq\{\mathbf{x}_{t}^{\prime},\mathbf{m}_{t}\}$.
The denoising step for $\mathbf{x}_{}^{n-1}$ can be expressed as follows, based on \equationname\eqref{eq:ddpm_sampleing_step}:
\begin{align}
&\mathbf{x}_{t}^{n-1} \sim \mathcal{N}(\mathbf{x}^{n-1}_{t}; \boldsymbol{\mu}_{\theta}^{\text{guided}}(\mathbf{x}^{n}_{t}, n, \mathbf{c}_t), \mathbf{\Sigma}^{n}).
\end{align}
Here,
$
\boldsymbol{\mu}_{\theta}^{\text{guided}}(\mathbf{x}_{t}^n, n, \mathbf{c}_t) :=
\frac{1}{\sqrt{\alpha_n}} (\mathbf{x}_{t}^n - \frac{1 - \alpha_n}{\sqrt{1 - \bar{\alpha}_n}} \cdot \epsilon_\theta^{\text{guided}}),
$ and thus $\epsilon_{\theta}^{\text{guided}}$ is computed as follows according to \equationname\eqref{eq:cfg_epsilon_guidance}.
\begin{equation}
\epsilon_\theta^{\text{guided}} = (1 + w)\epsilon_{\theta}(\mathbf{x}^{n}_{t}, n, \mathbf{c}_t) - w\epsilon_{\theta}(\mathbf{x}^{n}_{t}, n).
\end{equation}

\subsection{Cutting Action Planning}\label{subsection:cutting_planning}
As requirements for planning the next cutting action, it is desirable to (i) avoid damage to the target internal part and (ii) maximize the removable volume.
Given the presence score map $\mathbf{c}_{t}^{\text{map}}$, which represents an estimate of the target part's presence on each cutting surface based on the predicted internal structures $\mathcal{X}_{t}$, one possible formulation for planning the next cutting action $\mathbf{a}_{t+1}$ is given by
\begin{equation}
\mathbf{a}_{t+1}^{}
 = \argmax_{\mathbf{a} \in \mathcal{A}_{\text{feasible}}(\mathbf{c}_{t}^{\text{map}})}
    f_{\mathrm{vol}}(\mathbf{a}).
\label{eq:objective_function}
\end{equation}
Here, $\mathcal{A}_{\text{feasible}}(\cdot)$ is the set of candidate cutting actions for which the presence score map $\mathbf{c}_{t}^{\text{map}}$ does not exceed the cutting-risk threshold $\eta\ (\eta \ge 0)$:
\begin{equation}
\mathcal{A}_{\text{feasible}}(\mathbf{c}_{t}^{\text{map}})
  = \{ \mathbf{a} \in \mathcal{A} \mid \mathbf{c}_{t}^{\text{map}}(i,j) \le \eta \}.
\end{equation}
$f_{\mathrm{vol}}(\cdot)$ is a function that computes the removal volume corresponding to a cutting position.
Therefore, a small $\eta$ reduces the risk of damaging the target part but limits the removal volume, whereas a large $\eta$ increases this risk but allows larger removal volume.

\begin{figure}
        \centering
        \includegraphics[clip, width=1.0 \columnwidth]{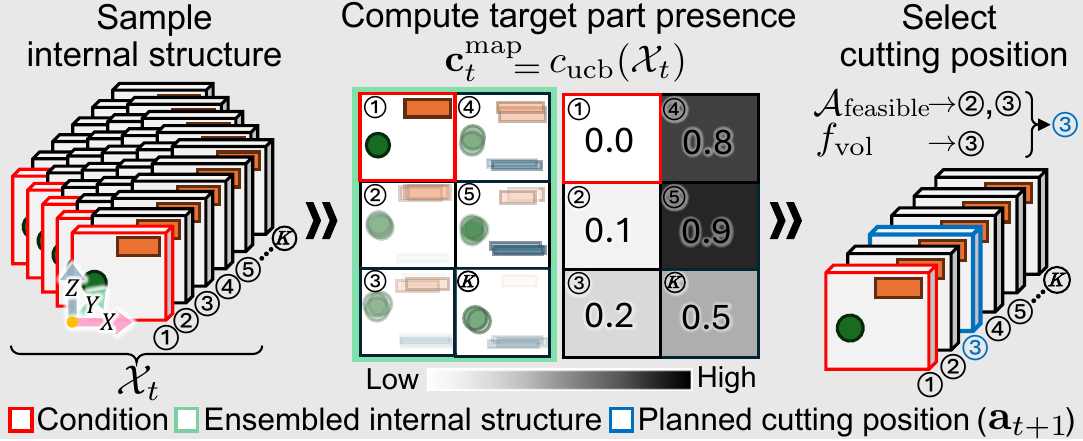}
        \caption{
        Cutting action planning along the $Y$-axis based on predicted internal structure.
        Values on each cutting surface indicate the presence score of the target internal part.
        Positions {\scriptsize{\Circled{2}}} and {\scriptsize{\Circled{3}}} are feasible owing to their low presence scores,
        and {\scriptsize{\Circled{3}}} is selected because it maximizes the removable volume.}
        \label{fig:cutting_plan_method}
\end{figure}

In our method, $\mathbf{c}_{t}^{\text{map}}$ is defined as the mean and standard deviation of $f_s(\mathcal{X}_t)$, where $f_s(\cdot)$ is the part detector, i.e., a binary classifier that indicates whether the target voxel feature $\mathbf{f}^{\text{target}}$ exists on each cutting surface $\{\mathbf{x}_t^{(i,j)}\}^m$:
\begin{equation}
\begin{aligned}
\mathbf{c}_{t}^{\text{map}}(i,j)&= c_{\text{ucb}}(\mathcal{X}_{t}; i,j),\\
&= \boldsymbol{\mu}_{m}\!(f_{s}(\mathbf{x}_{t}^{(m)}; i,j))+ \gamma \, \boldsymbol{\sigma}_{m}\!(f_{s}(\mathbf{x}_{t}^{(m)}; i,j)),
\end{aligned}
\label{eq:c_ucb}
\end{equation}
where $\boldsymbol{\mu}_{m}(\cdot)$ and $\boldsymbol{\sigma}_{m}(\cdot)$ compute the mean and standard deviation, respectively, and $\gamma$ is the weight for the standard deviation\footnote{While we use a UCB-style score $\mu + \gamma\sigma$ in Eq.~(11), the planner is not tied to this particular choice. Any risk function over the Monte Carlo samples ${f_s(\cdot)}$ can be used to define $\mathbf{c}_{t}^{\text{map}}$, e.g., a Value at Risk~\cite{RiskMetrics1996}, Conditional Value at Risk~\cite{Stan_2000_CVaR}, or a worst-case (max-sample) score. These alternatives trade off conservativeness and efficiency in different ways and can be selected depending on application requirements.}.
Part detector $f_s(\cdot)$ is defined as
\begin{equation}
f_{s}(\mathbf{x}_{t}; i,j) = \mathds{1} [\mathbf{f}^{\text{target}} \in \mathbf{x}_{t}^{(i,j)}].
\label{eq:part_detector_func}
\end{equation}
In this work, each cutting surface is represented as a set of voxels.
Therefore, the target part's presence is evaluated by determining if
any voxel in the cutting surface corresponds to the target voxel feature $\mathbf{f}^{\text{target}}$.
Figure~\ref{fig:cutting_plan_method} illustrates how the cutting action is planned along the $Y$-axis.

\section{Experiments}
To verify the effectiveness of the proposed method, we built a geometric cutting simulator and conducted experiments on simple-shaped models composed of primitive geometries
and complex-shaped models that emulate real products.
This experiment had two objectives:
\begin{itemize}
    \item To confirm that the proposed method enables the extraction of a target internal part based on the observed cutting surfaces.
    In particular, we evaluate the effect of the cutting-risk threshold $\eta$, used for cutting planning (\S \ref{paragrah:simple_costmap_eval}, \ref{paragrah:complex_costmap_eval}).
    \item To compare the task performance of the proposed method with those of baseline methods (\S \ref{paragrah:simple_pref_eval}, \ref{paragrah:complex_pref_eval}).
\end{itemize}

The iterative cutting process for both the simple- and complex-shaped models is shown in Video~1 (01:38--04:13).

\subsection{Common Experimental Setup}
\subsubsection{Preparation of Benchmark Models}
As shown in \figurename\ref{fig:bench_mark_shape}, the simple-shaped models consist of an external shape (gray) enclosing three cuboids, while the complex-shaped models emulate a real product (a sheet sander) with three components: a battery (blue), a PCB (green), and a motor (red).
In the simple-shaped models, Part~B was set as the battery and extracted as the target internal part.
In these experiments, we assumed that the color of each part could be identified from the observed cutting surfaces,
and we used the identified color as voxel feature $\mathbf{f}_k\coloneqq [R,G,B] \in \mathbb{R}^3$.
In real products, analogous part-related cues (which we represent here as color-coded voxel features) can be obtained from raw cutting-surface observations by leveraging modern segmentation models (e.g., Mask R-CNN~\cite{Kaiming_2017_maskrcnn} or SAM~\cite{Kirillov_2023_ICCV,Ravi_2025_sam2,carion_2026_sam3}); see \S\ref{sec:discussion} for further discussion.
\subsubsection{Simulation Environments}
For shape editing and rendering, we extended PyVista \cite{sullivan2019pyvista} and Blender \cite{blender}.
While cutting errors and observation noise can occur in real environments,
the simulation experiments assumed planned cuts were executed without error and cutting surfaces were observed accurately. This setting focuses on evaluating the core capability of the proposed framework to estimate internal structures from observed cutting surfaces and to extract the target part by leveraging the estimated uncertainty.
At each step, unnecessary regions were assumed to be discarded immediately after cutting and thus became unobservable thereafter.
\subsubsection{Baseline Methods}\label{subsection:comparison_method}
We prepared the following baseline methods.
\begin{itemize}
    \item \textbf{Random}: A method that randomly selects cutting actions at each task step and performs cutting.
    \item \textbf{VAEAC}: A method that predicts internal structures using a CVAE-based generative model \cite{ivanov2018variational}.
    \item \textbf{PCD-DM}: A method that directly predicts 3D shapes represented as colored point clouds using a diffusion model with a 1D-UNet.
    While the proposed method outputs $3K^3$ dimensions for voxel colors, \textbf{PCD-DM} directly outputs the position $\mathbf{p}$ and color $\mathbf{f}$ of each point, with an output dimensionality of $6K^3$.
    \item \textbf{Proposed-Nocond}: The proposed method without conditioning on the observed cutting surfaces.
    \item \textbf{Proposed-GT}:
    A method that plans cuts with known internal structures, which is used to evaluate the upper bound of task performance for the proposed method.
\end{itemize}
Except for \textbf{Random}, all baseline methods used the same cutting action planning procedure as the proposed method.
\subsubsection{Task Performance Metrics}
We prepared three metrics to evaluate task performance, where the volumes were computed from the number of voxels.
\begin{itemize}
    \item \textbf{Cutting Error Volume} [voxels]:
     Total volume of the target part that was cut erroneously during the task.
    \item \textbf{Part Remaining Rate} [\%]: Retention rate of the target part, defined as
    (volume of target part after task/volume of target part before task) $\times 100$.
    \item \textbf{Part Occupancy Rate} [\%]: Proportion of the target part volume to the total volume of the shape after the task, defined as (volume of target part after task/total volume of shape after task) $\times 100$.
\end{itemize}
\subsubsection{Cutting Action Planning Settings}
In this experiment, since color information was used as the voxel feature, we computed the part detector $f_s(\mathcal{X}_{t})$ given by \equationname\eqref{eq:part_detector_func} for each cutting surface $\mathbf{x}_{t}^{(i,j)}$ to evaluate the target part's presence, as follows:
\begin{equation}
f_{s}(\mathbf{x}_{t}; i,j) = \mathds{1}[\exists d \;\text{s.t.}\;\;
   \mathbf{f}_{\min}^{\text{target}} \leq \mathbf{x}_{t}^{(i,j)}[d] \leq \mathbf{f}_{\max}^{\text{target}}],
\end{equation}
where, $\mathbf{x}_{t}^{(i,j)}[d]$ denotes the color value of the \(d\)-th voxel on the cutting surface $d \in \{1, \dots, K^2\}$ and $\mathbf{f}_{\min}^{\text{target}},\ \mathbf{f}_{\max}^{\text {target}}$ represent the range of color values of the target part.

In \equationname\eqref{eq:c_ucb}, we set the scale parameter \(\gamma=1.0\)
to equally evaluate the mean and the standard deviation. As a simple implementation, we solve the optimization in \equationname\eqref{eq:objective_function} via an exhaustive search over all cutting surfaces.
The number of task steps \(T\) was set to 8.
\subsection{Experimental Setup of Simple-Shaped Model}
\begin{figure}
    \centering
    \includegraphics[clip, width=1.0\linewidth]{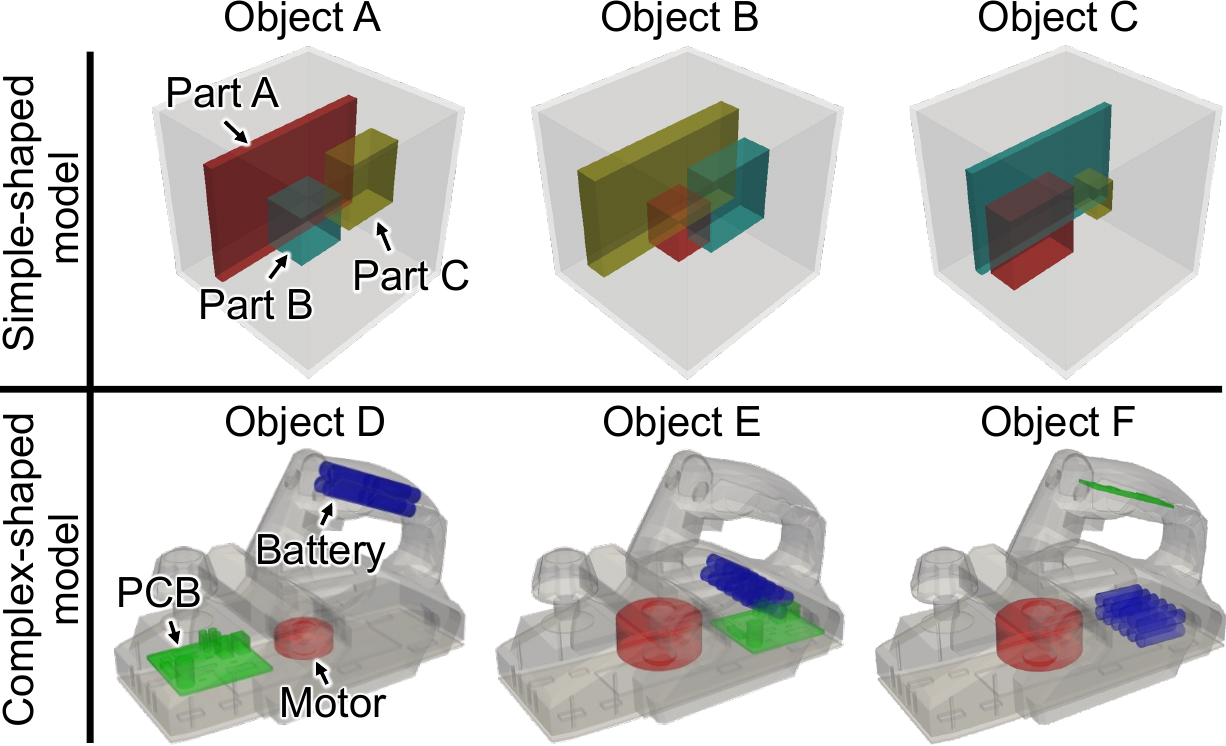}
    \caption{
    Object configurations used for evaluation.
    The battery is defined as the target internal part for extraction.
    In the simple-shaped model, Part B corresponds to the battery. }
    \vspace{-4truemm} 
    \label{fig:bench_mark_shape}
\end{figure}

We prepared five internal-part arrangement distributions and generated 20K samples from each to train the conditional diffusion model.
Each distribution had different part sizes and relative spatial relationships (reference arrangements), and the final placement was sampled from a uniform distribution centered on the reference arrangement.
The internal structures used for evaluation (\figurename\ref{fig:bench_mark_shape}) were newly sampled from the same three arrangement distributions used in training.
This evaluation follows a standard independent and identically distributed (i.i.d.) setting in statistical machine learning \cite{bishop2006prml}, and aims to validate whether the learned model can represent the multi-modal internal-part arrangements of the training distribution.

3D shapes were voxelized at $K = 16$ and used to train the diffusion model with a U-Net architecture for 100K steps (batch size 192, $N = 1000$ diffusion steps).
During training, the diffusion model is conditioned on randomly generated masks~\cite{ivanov2018variational,Pathak2016ContextEF} to cover diverse observation patterns of cutting surfaces.
Sampling was performed with a guidance scale of $w=0.2$ using $M=32$ samples.
Training the diffusion model took approximately $16.4$ hours on NVIDIA GeForce RTX 4090. At inference time, we used DDIM sampling~\cite{song_2021_ddim} with 20 denoising steps to accelerate generation; on average, producing $M=32$ conditional samples for one planning step took $0.7$ seconds on the same hardware.

\begin{figure*}
        \centering
        \includegraphics[clip, width=0.99\linewidth]{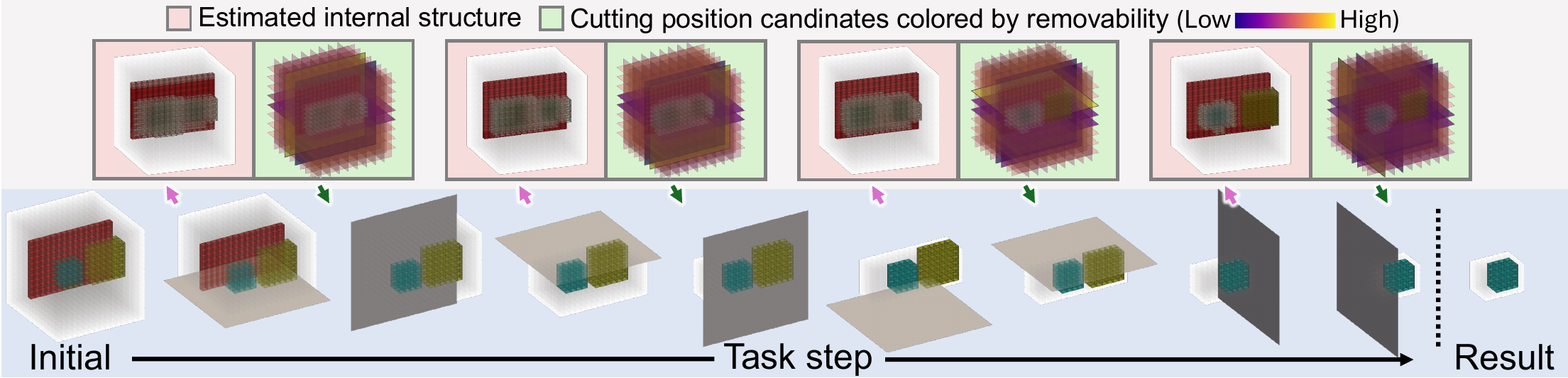}
        \caption{
        Examples of cutting process transitions by the proposed VoxelDiffusionCut with the simple-shaped model (Object A).
        VoxelDiffusionCut plans cutting positions based on the estimated internal structure to maximize the removable volume while non-destructively extracting the target internal part.}
        \vspace{-4truemm} 
        \label{fig:sim:cutting_process}
\end{figure*}

\begin{table}
\begin{minipage}{\linewidth}
\caption{
Task performance comparison under different cutting-risk thresholds for the simple-shaped model (Object A)}
\label{tab:sim:simple_model_costmap_comp_object_6}
\centering
\setlength{\tabcolsep}{3.0pt}
\scalebox{0.95}{
\begin{tabular}{ccccc}
\toprule
\( \eta \) & Cut Err. Vol. $\downarrow$ & Part Remain. Rate $\uparrow$ & Part Occ. Rate $\uparrow$ \\
\midrule
0.0& \textbf{0.0} $\pm$0.0 & \textbf{100.0} $\pm$0.0 &  7.71 $\pm$0.57\\
0.5& \textbf{0.0} $\pm$0.0 & \textbf{100.0} $\pm$0.0 &  \textbf{27.17} $\pm$2.66\\
1.0& 188.0 $\pm$78.59 & 25.0 $\pm$28.87 & 2.84 $\pm$3.33\\
\bottomrule
\end{tabular}}
\vspace{2.3truemm}
\end{minipage}
\begin{minipage}{\linewidth}
\caption{Task performance comparison of simple-shaped models.
In the part occupancy rate column, boldface figures indicate the highest value among the methods without cutting error volume.
}
\label{tab:sim:task_performance_comparison_all}
\centering
\begin{subtable}{\linewidth}
\centering
\caption{Object A}
\label{tab:sim:task_performance_comparison_object_A}
\setlength{\tabcolsep}{2.0pt}
\scalebox{0.805}{
\begin{tabular}{lcccc}
\toprule
Method & Cut Err. Vol. $\downarrow$ & Part Remain. Rate $\uparrow$  & Part Occ. Rate $\uparrow$ \\
\midrule
Random          & 148.67 $\pm$106.57 & 42.36 $\pm$21.57 & 2.5 $\pm$1.16\\
VAEAC           & 84.0 $\pm$0.0 & 75.0 $\pm$0.0 & 60.0 $\pm$0.0\\
PCD-DM    & 66.0 $\pm$30.0 & 50.0 $\pm$25.0 & 42.25 $\pm$8.7\\
Proposed-Nocond & \textbf{0.0} $\pm$0.0 & \textbf{100.0} $\pm$0.0 & 8.88 $\pm$1.56\\
Proposed        & \textbf{0.0} $\pm$0.0 & \textbf{100.0} $\pm$0.0 & \textbf{27.17} $\pm$2.66\\
\hdashline
Proposed-GT     & 0.0 $\pm$0.0 & 100.0 $\pm$0.0 & 48.0 $\pm$0.0\\
\bottomrule
\end{tabular}
}
\end{subtable}
\\[0.5em]
\begin{subtable}{\linewidth}
\centering
\caption{Object B}
\label{tab:sim:task_performance_comparison_object_B}
\setlength{\tabcolsep}{2.0pt}
\scalebox{0.805}{
\begin{tabular}{lcccc}
\toprule
Method & Cut Err. Vol. $\downarrow$ & Part Remain. Rate $\uparrow$  & Part Occ. Rate $\uparrow$ \\
\midrule
Random          & 490.5 $\pm$157.2 & 30.69 $\pm$15.54 & 4.59 $\pm$2.54\\
VAEAC           & 252.0 $\pm$0.0 & 71.43 $\pm$0.0 & 75.0 $\pm$0.0\\
PCD-DM    & 304.5 $\pm$109.62 & 41.67 $\pm$20.97 & 51.53 $\pm$5.24\\
Proposed-Nocond & \textbf{0.0} $\pm$0.0 & \textbf{100.0} $\pm$0.0 & 21.87 $\pm$1.4\\
Proposed        & \textbf{0.0} $\pm$0.0 & \textbf{100.0} $\pm$0.0 & \textbf{48.61} $\pm$4.35\\
\hdashline
Proposed-GT     & 0.0 $\pm$0.0 & 100.0 $\pm$0.0 & 58.33 $\pm$0.0\\
\bottomrule
\end{tabular}
}
\end{subtable}
\\[0.5em]
\begin{subtable}{\linewidth}
\centering
\caption{Object C}
\label{tab:sim:task_performance_comparison_object_C}
\setlength{\tabcolsep}{2pt}
\scalebox{0.805}{
\begin{tabular}{lcccc}
\toprule
Method & Cut Err. Vol. $\downarrow$ & Part Remain. Rate $\uparrow$  & Part Occ. Rate $\uparrow$ \\
\midrule
Random          & 558.67 $\pm$175.63 & 21.07 $\pm$15.92 & 5.75 $\pm$4.73\\
VAEAC           & 35.0 $\pm$10.69 & 90.0 $\pm$0.0 & 21.87 $\pm$3.09\\
PCD-DM    & 291.67 $\pm$103.01 & 65.0 $\pm$11.18 & 31.35 $\pm$12.48\\
Proposed-Nocond & 767.0 $\pm$309.49 & 0.0 $\pm$0.0 & 0.0 $\pm$0.0\\
Proposed        & \textbf{0.0} $\pm$0.0 & \textbf{100.0} $\pm$0.0 & \textbf{26.52} $\pm$0.0\\
\hdashline
Proposed-GT     & 0.0 $\pm$0.0 & 100.0 $\pm$0.0 & 39.77 $\pm$0.0\\
\bottomrule
\end{tabular}
}
\end{subtable}
\end{minipage}
\end{table}

\subsection{Results of Simple-shaped Models}\label{sec:simple_model_exp}
\subsubsection{{Task Performance Comparison with Different Cutting-Risk Thresholds}}\label{paragrah:simple_costmap_eval}
TABLE~\ref{tab:sim:simple_model_costmap_comp_object_6} shows the task performance of Object~A evaluated six times with different cutting-risk thresholds $\eta$.
The results indicate that with $\eta=0.0$, erroneous cuts are completely avoided, but the part occupancy rate is low, reflecting a conservative cutting plan.
With $\eta=1.0$, the part occupancy rate is high, but the cutting error volume is large.
With $\eta=0.5$, erroneous cuts are avoided while maintaining a high part occupancy rate.
Figure~\ref{fig:sim:cutting_process} shows the cutting process of Object~A ($\eta=0.5$), which was planned based on the estimated internal structure.

These results show that the proposed method achieves efficient removal of unnecessary regions without damaging the target part through the adjustment of $\eta$.
In the following experiments, we used \(\eta = 0.5\), which produced a balanced performance.

\begin{figure}
    \begin{minipage}[t]{\columnwidth}
        \centering
        \includegraphics[width=1.0\columnwidth]{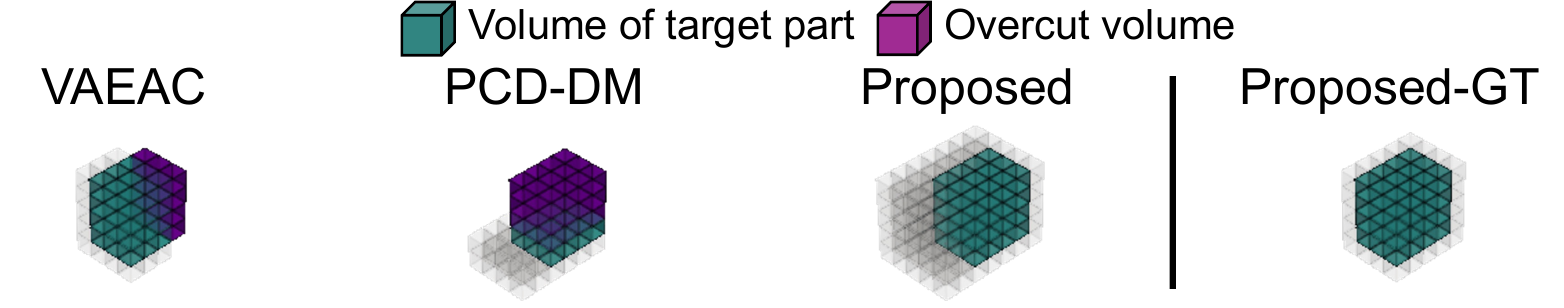}
        \\[0.5em]
        \subcaption{Shape examples of Object A}
        \label{fig:sim:Object_A_image_render}
    \end{minipage}
    \begin{minipage}[t]{\columnwidth}
        \centering
        \vspace{1.5truemm}
        \includegraphics[clip,width=1.0\columnwidth]{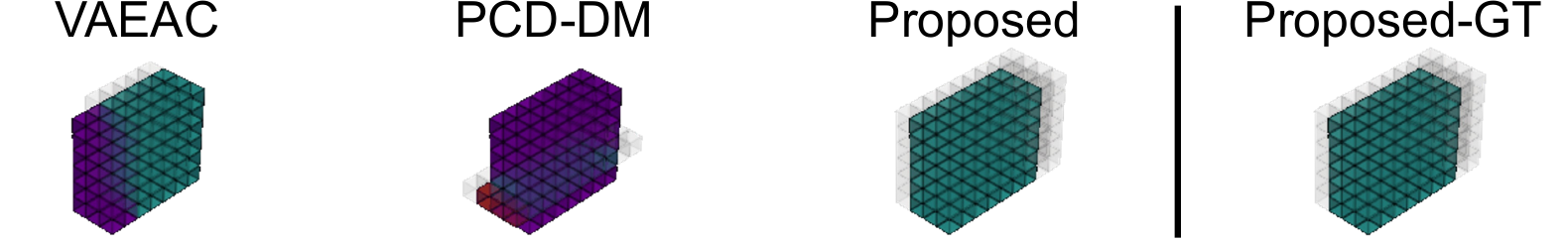}
        \\[0.5em]
        \subcaption{Shape examples of Object B}
        \label{fig:sim:Object_B_image_render}
    \end{minipage}
    \begin{minipage}[t]{\columnwidth}
        \centering
        \vspace{1.5truemm}
        \includegraphics[clip,width=1.0\columnwidth]{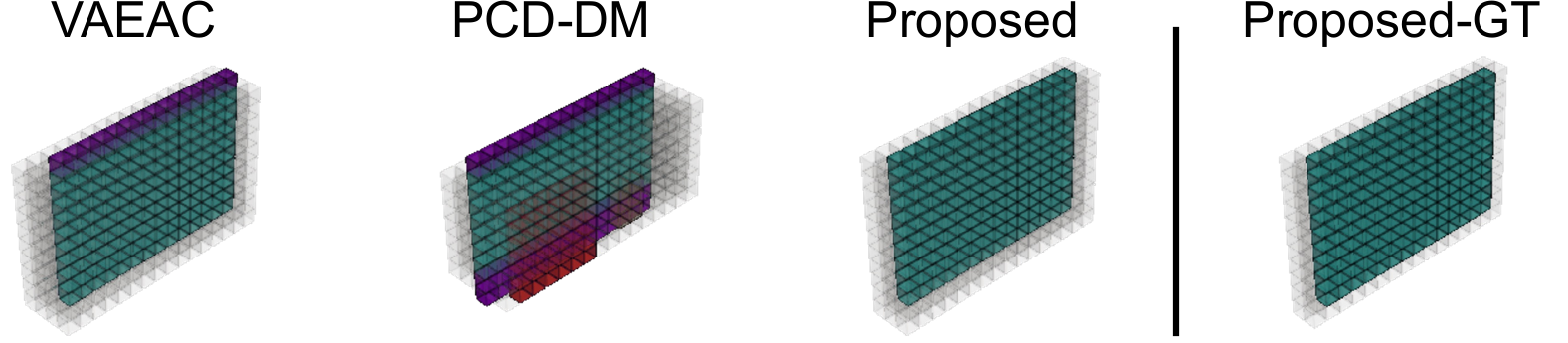}
        \\[0.5em]
        \subcaption{Shape examples of Object C}
        \label{fig:sim:Object_C_image_render}
    \end{minipage}
    \caption{Rendered objects after processing by each method for the simple-shaped models}
    \label{fig:sim_result:simple_object_all}
\end{figure}

\subsubsection{Task Performance Comparison with Baseline Methods}\label{paragrah:simple_pref_eval}
TABLE~\ref{tab:sim:task_performance_comparison_all} shows the task performance of each method evaluated six times.
Figure~\ref{fig:sim_result:simple_object_all} visualizes examples of the extracted shapes.
\textbf{VAEAC} and \textbf{PCD-DM} suffer from erroneous cuts, leading to a low part remaining rate. At the same time, these erroneous cuts explain why their part occupancy rate occasionally appears high: Cutting into the target part increased the observed information on the internal structure and thus improved inference accuracy.
Compared with \textbf{Proposed-Nocond}, which is not conditioned on the observed cutting surfaces, the proposed method achieves a consistently higher part occupancy rate by explicitly conditioning on them.

These results indicate that, for multiple evaluated shapes with different internal structure configurations, the proposed method produces cutting plans that reduce erroneous cuts of the target part while maintaining a higher part occupancy rate than the baseline approaches.

\begin{figure*}
        \centering
        \includegraphics[clip, width=0.99\linewidth]{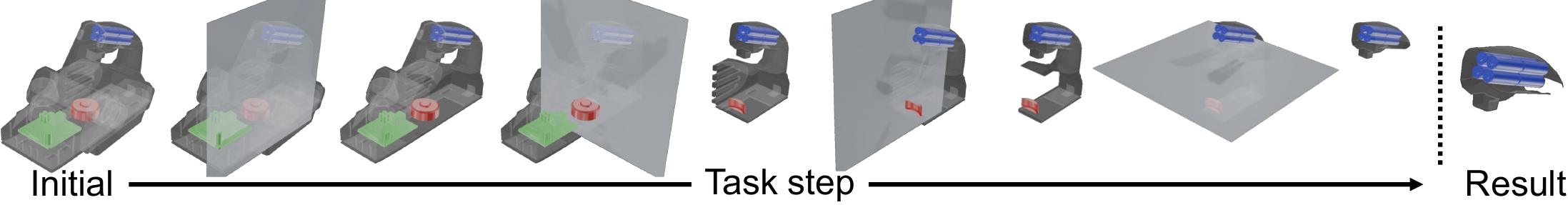}
        \caption{Examples of cutting process transitions by the proposed VoxelDiffusionCut with the complex-shaped model (Object D)}
        \label{fig:sim:complex_cutting_process}
\end{figure*}

\begin{table}[t]
    \begin{minipage}{\linewidth}
        \caption{Task performance comparison under different cutting-risk thresholds
        for the complex-shaped model (Object D)}
        \label{tab:sim:complex_model_costmap_comp}
        \centering
        \setlength{\tabcolsep}{3.0pt}
        \scalebox{0.95}{
        \begin{tabular}{ccccc}
        \toprule
        \(
         \eta
        \) & Cut Err. Vol. $\downarrow$ &  Part Remain. Rate $\uparrow$  & Part Occ. Rate $\uparrow$ \\
        \midrule
        0.0& \textbf{0.0} $\pm$0.0 & \textbf{100.0} $\pm$0.0 & 3.94 $\pm$2.14\\
        0.5& \textbf{0.0} $\pm$0.0 & \textbf{100.0} $\pm$0.0 & \textbf{33.14} $\pm$6.25\\
        1.0& 38.33 $\pm$55.81 & 88.89 $\pm$15.71 & 20.0 $\pm$8.26\\
        \bottomrule
        \end{tabular}
        }
        \vspace{2.3truemm}
    \end{minipage}
    \begin{minipage}{\linewidth}
        \caption{Task performance comparison of complex-shaped models.
        In the part occupancy rate column, boldface figures indicate the highest value
        among the methods without cutting error volume.
        }
        \label{tab:sim:complex_task_performance_comparison_all}
        \centering
        \begin{subtable}{\linewidth}
        \centering
        \caption{Object D}
        \label{tab:sim:complex_task_performance_comparison_object_A}
        \setlength{\tabcolsep}{2pt}
        \scalebox{0.805}{
        \begin{tabular}{lcccc}
        \toprule
        Method & Cut Err. Vol. $\downarrow$ & Part Remain. Rate $\uparrow$  & Part Occ. Rate $\uparrow$ \\
        \midrule
        Random          & 332.17 $\pm$252.73 & 57.0 $\pm$25.49 & 0.19 $\pm$0.09\\
        VAEAC           & \textbf{0.0} $\pm$0.0 & \textbf{100.0} $\pm$0.0 & 0.24 $\pm$0.0 \\
        PCD-DM          & \textbf{0.0} $\pm$0.0 & \textbf{100.0} $\pm$0.0 & 0.24 $\pm$0.0 \\
        Proposed-Nocond & \textbf{0.0} $\pm$0.0 & \textbf{100.0} $\pm$0.0 & 4.3 $\pm$0.68\\
        Proposed        & \textbf{0.0} $\pm$0.0 & \textbf{100.0} $\pm$0.0 & \textbf{33.14} $\pm$6.25\\
        \hdashline
        Proposed-GT     & 0.0 $\pm$0.0 & 100.0 $\pm$0.0 & 35.94 $\pm$0.0\\
        \bottomrule
        \end{tabular}
        }
        \end{subtable}
        \\[0.5em]
        \begin{subtable}{\linewidth}
        \centering
        \caption{Object E}
        \label{tab:sim:complex_task_performance_comparison_object_B}
        \setlength{\tabcolsep}{2pt}
        \scalebox{0.805}{
        \begin{tabular}{lcccc}
        \toprule
        Method & Cut Err. Vol. $\downarrow$ & Part Remain. Rate $\uparrow$  & Part Occ. Rate $\uparrow$ \\
        \midrule
        Random          & 275.33 $\pm$128.77 & 70.71 $\pm$13.66 & 0.43 $\pm$0.09\\
        VAEAC           & 25.0 $\pm$0.0 & 89.18 $\pm$0.0 & 61.31 $\pm$0.0 \\
        PCD-DM          & \textbf{0.0} $\pm$0.0 & \textbf{100.0} $\pm$0.0 & 0.41 $\pm$0.0\\
        Proposed-Nocond & 26.0 $\pm$58.14 & 94.37 $\pm$12.58 & 5.4 $\pm$1.52\\
        Proposed        & \textbf{0.0} $\pm$0.0 & \textbf{100.0} $\pm$0.0 & \textbf{41.39} $\pm$9.03\\
        \hdashline
        Proposed-GT     & 0.0 $\pm$0.0 & 100.0 $\pm$0.0 & 55.0 $\pm$0.0\\
        \bottomrule
        \end{tabular}
        }
        \end{subtable}
        \\[0.5em]
        \begin{subtable}{\linewidth}
        \centering
        \caption{Object F}
        \label{tab:sim:complex_task_performance_comparison_object_C}
        \setlength{\tabcolsep}{2pt}
        \scalebox{0.805}{
        \begin{tabular}{lcccc}
        \toprule
        Method & Cut Err. Vol. $\downarrow$ & Part Remain. Rate $\uparrow$  & Part Occ. Rate $\uparrow$ \\
        \midrule
        Random          & 261.5 $\pm$68.83 & 57.78 $\pm$8.64 & 0.2 $\pm$0.03\\
        VAEAC           & 75.0 $\pm$0.0 & 88.89 $\pm$0.0 & 23.44 $\pm$0.0\\
        PCD-DM          & \textbf{0.0} $\pm$0.0 & \textbf{100.0} $\pm$0.0 & 0.25 $\pm$0.0\\
        Proposed-Nocond & \textbf{0.0} $\pm$0.0 & \textbf{100.0} $\pm$0.0 & 3.35 $\pm$0.96\\
        Proposed        & \textbf{0.0} $\pm$0.0 & \textbf{100.0} $\pm$0.0 & \textbf{5.47} $\pm$0.0\\
        \hdashline
        Proposed-GT     & 0.0 $\pm$0.0 & 100.0 $\pm$0.0 & 56.25 $\pm$0.0\\
        \bottomrule
        \end{tabular}
        }
        \end{subtable}
    \end{minipage}
\end{table}

\begin{figure}
\begin{minipage}[t]{\columnwidth}
    \centering
        \includegraphics[clip,width=1.0\columnwidth]{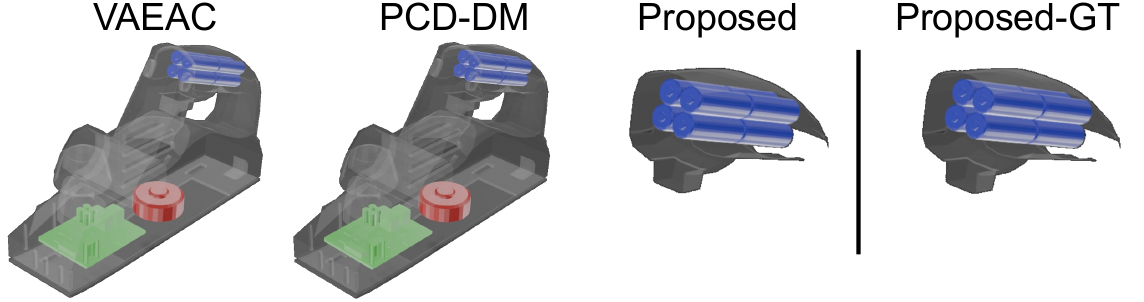}
    \\[0.5em]
    \subcaption{Shape examples of Object D}
    \label{fig:sim_result:Object_A_image_render}
\end{minipage}
\begin{minipage}[t]{\columnwidth}
    \centering
    \vspace{1.5truemm}
        \includegraphics[clip,width=1.0\columnwidth]{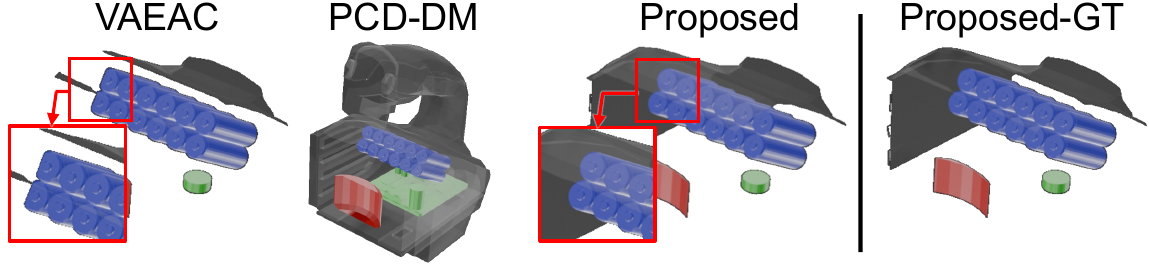}
    \\[0.5em]
    \subcaption{Shape examples of Object E}
    \label{fig:sim_result:Object_B_image_render}
\end{minipage}
\begin{minipage}[t]{\columnwidth}
    \centering
    \vspace{1.5truemm}
        \includegraphics[clip,width=1.0\columnwidth]{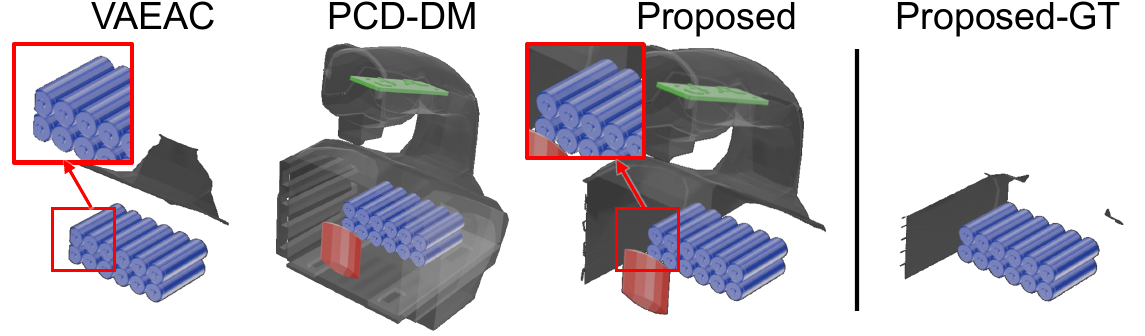}
        \\[0.5em]
        \subcaption{Shape examples of Object F}
    \label{fig:sim_result:Object_C_image_render}
\end{minipage}
\caption{Rendered objects after processing by each method for complex-shaped models. The areas enlarged in red frames show where the baseline method erroneously cut the target internal part.}
\vspace{-2truemm}
\label{fig:sim_result:Object_all}
\end{figure}

\subsection{Experimental Setup of Complex-shaped Models}\label{sec:complex_model_exp}
We prepared six internal-part arrangement distributions and generated 3K samples from each to train the conditional diffusion model.
Each distribution had different part sizes and relative spatial relationships, as with the simple-shaped models.
The final placement was sampled from a uniform distribution centered on the reference arrangement.
The internal structures used for evaluation (\figurename\ref{fig:bench_mark_shape}) were newly sampled from the same arrangement distributions used in training, following the same evaluation setting as for the simple-shaped models.

3D shapes were voxelized at $K = 46$ and used to train the diffusion model with a U-Net architecture for 100K steps (batch size 90, $N = 1000$ diffusion steps).
During training, the diffusion model is conditioned on randomly generated masks~\cite{ivanov2018variational,Pathak2016ContextEF} to cover diverse observation patterns of cutting surfaces.
Sampling was performed with a guidance scale of $w=0.2$ using $M=32$ samples.
Training the diffusion model for 100K steps took approximately $5.3$ days on (NVIDIA H100 NVL).
Inference was performed with DDIM sampling~\cite{song_2021_ddim} as in the simple-shaped setup; producing $M=32$ conditional samples with 20 denoising steps for one planning step took $26.7$ seconds on average on the same hardware.

\subsection{Results of Complex-shaped Models}
\subsubsection{Task Performance Comparison with Different Cutting-Risk Thresholds}\label{paragrah:complex_costmap_eval}
TABLE~\ref{tab:sim:complex_model_costmap_comp} shows the task performance of Object D evaluated six times with different cutting-risk thresholds $\eta$.
These results show that the proposed method achieves efficient removal of unnecessary regions without damaging the target part through the adjustment of $\eta$, similar to the simple-shaped models.
Figure~\ref{fig:sim:complex_cutting_process} visualizes the cutting process of Object D ($\eta=0.5$).
\subsubsection{Task Performance Comparison with Baseline Methods}\label{paragrah:complex_pref_eval}
TABLE~\ref{tab:sim:complex_task_performance_comparison_all} shows the task performance of each method evaluated six times.
Figure~\ref{fig:sim_result:Object_all} visualizes examples of the extracted shapes.
\textbf{PCD-DM} avoids erroneous cuts but exhibits a low part occupancy rate. This suggests that \textbf{PCD-DM}, which directly generates 3D shapes as colored point clouds, struggles to plan effective cutting actions for extracting the target part based on the estimated internal structures in high-dimensional and complex shapes.

\begin{figure*}
\centering
\includegraphics[clip,width=1.0
\linewidth]{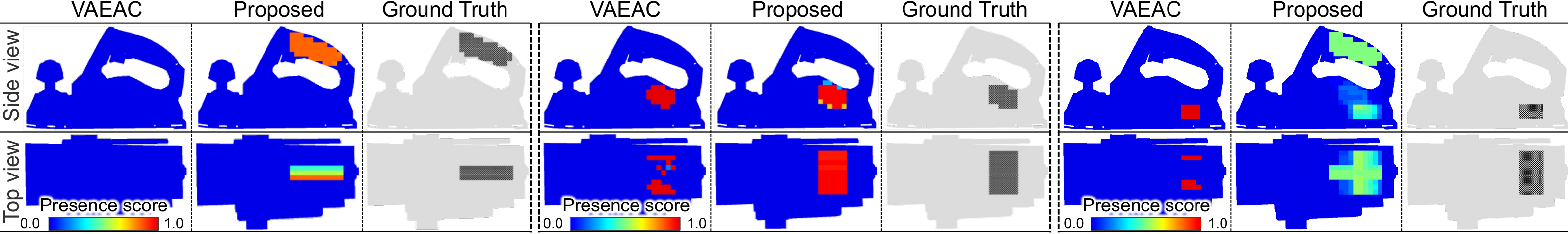}
\begin{minipage}[t]{0.32\linewidth}
    \centering
    \vspace{-3truemm}
    \subcaption{Object D}
    \label{fig:sim:complex_model_location_extimation_Object_D}
\end{minipage}
\begin{minipage}[t]{0.32\linewidth}
    \centering
    \vspace{-3truemm}
    \subcaption{Object E}
    \label{fig:sim:complex_model_location_extimation_Object_E}
\end{minipage}
\begin{minipage}[t]{0.32\linewidth}
    \centering
    \vspace{-3truemm}
    \subcaption{Object F}
    \label{fig:sim:complex_model_location_extimation_Object_C}
\end{minipage}
\vspace{-1.5truemm}
\caption{Comparison of presence score maps for target internal part locations in complex-shaped models at task step \(t\)=8
}
\label{fig:sim:complex_model_location_extimation_Object_all}
\end{figure*}

In Object~F, \textbf{VAEAC} produced erroneous cuts,
whereas \textbf{Proposed} avoided them but did not achieve battery-only extraction.
Figure~\ref{fig:sim:complex_model_location_extimation_Object_all} shows the presence score map of the estimated internal structure at task step \(t=8\).
The figure shows that the proposed method, which uses a conditional diffusion model, provided high-confidence internal structure estimation for Objects D and E while capturing uncertainty for Object F. In contrast, \textbf{VAEAC} had errors in estimating the internal structure across all objects.
These results indicate that the proposed method provides uncertainty-aware estimation and cutting plans, even when the internal structure cannot be represented as a single distribution from the observed cutting surfaces. This characteristic is essential at recycling and disposal sites, where parts such as batteries must be extracted without damage for safety.

These results confirm that, even for high-dimensional, complex-shaped models, the proposed method provides cutting plans that reduce erroneous cuts under high uncertainty and achieve a high part occupancy rate when the internal structure can be estimated.

\section{Discussion}\label{sec:discussion}
\subsection{Initial Cutting Selection}\label{sec:discussion:init}
A limitation of the proposed method is the need to predefine the cutting position at the initial step.
Since our method estimates internal structure from observed cutting surfaces, it cannot be applied when such observations are unavailable.
To address this, we are investigating an extension that leverages external shape and cutting-surface observations through conditioning on information such as product type and appearance.
This extension is expected to permit automatic selection of the initial cutting position from external information.

\subsection{Real-World Deployment}\label{sec:discussion:real}
Applying the proposed method to real environments is an interesting direction for future work.
One additional challenge in such settings is the observation noise of cutting surfaces, e.g., irregular textures, debris, occlusions, or illumination changes.
Recent advances in object recognition and segmentation models have demonstrated robustness to noisy observations, ranging from conventional instance segmentation models such as Mask R-CNN~\cite{Kaiming_2017_maskrcnn} to more recent foundation models such as the Segment Anything Model (SAM)~\cite{Kirillov_2023_ICCV,Ravi_2025_sam2,carion_2026_sam3}.
Therefore, incorporating such models to identify internal part types from observed cutting surfaces may enable the proposed loop to operate under realistic visual variability, by using generic cues such as geometry, texture, and material-dependent appearance instead of assuming pre-defined color labels.

Another challenge is related to the execution of planned cutting actions.
Physical constraints, including fixed product posture, limited manipulator workspace, and deviations caused by cutting resistance forces, can lead to discrepancies between planned and executed cuts. 
Such discrepancies are particularly relevant for contact-rich cutting tools (e.g., saw- or grinding-based operations), where tool deflection and reaction forces may introduce large cutting errors.
In grinding operations, automation methods have been proposed to compensate for errors due to grinding resistance~\cite{10214100}.
Therefore, it is important to plan cutting positions that consider both the estimated internal structure and such physical constraints.

At the same time, it is worth noting that the idealized assumptions used in this paper are not necessarily far from certain modern industrial cutting processes.
For example, abrasive waterjet cutting~\cite{Yang_2021_waterjet} can produce relatively clean and planar cuts under appropriate fixturing and process conditions, reducing surface irregularities and geometric deviations compared to contact-rich cutting tools.
In addition, diamond wire sawing~\cite{Deng_2022_DWS} has been reported as an effective separation process in photovoltaic module recycling, enabling high-precision cutting with reduced kerf loss under controlled conditions.
In such settings, the gap between the nominal cut plane and the realized cut, as well as the ambiguity in the observed cutting surface, can be substantially smaller, making our simulation setting a reasonable first approximation.
A systematic evaluation under realistic noise and execution-error models, together with process-specific constraints and cost considerations, remains an important direction for future work.

\section{Conclusion}
We propose VoxelDiffusionCut, a framework for non-destructive extraction of the target internal part from a product with an unknown internal structure using observations on the cutting surfaces.
The problem was formulated as a generative task, in which a diffusion model was applied to complete the voxel representation based on the observed cutting surfaces.
Experimental results in simulation suggest that the proposed method can estimate internal
structures from observed cutting surfaces and enable non-destructive extraction of the target internal part by leveraging the estimated uncertainty.

\section{Acknowledgments}
This work is based on results obtained from a project, JPNP23002, commissioned by the New Energy and Industrial Technology Development Organization (NEDO).

\printcredits

\bibliographystyle{cas-model2-names}

\bibliography{reference}

\end{document}

%% file: setting.tex
\usepackage{orcidlink}

\usepackage{arydshln}
\usepackage{mathtools}
\usepackage{dsfont}
\usepackage{circledsteps}
\usepackage{caption}
\usepackage{subcaption}
\usepackage[labelsep=period]{caption} 

\renewcommand{\figurename}{Fig.~} 
\newcommand{\equationname}{Eq.~}
\newcommand{\argmax}{\mathop{\rm arg~max}\limits}

\newif\ifhighlight
\highlighttrue        